%% file: main.tex
\documentclass[10pt,twocolumn,letterpaper]{article}

\usepackage{cvpr}              %

\usepackage{graphicx}
\usepackage{amsmath}
\usepackage{amssymb}
\usepackage{booktabs}

\usepackage{epsfig}
\usepackage{graphicx}
\usepackage{xcolor}
\usepackage{amsmath}
\usepackage{amssymb}
\usepackage{subcaption}
\usepackage{textcomp}
\usepackage{gensymb}
\usepackage{mathtools}
\usepackage{amssymb}
\usepackage{wrapfig}
\usepackage{lipsum}
\usepackage{diagbox}
\usepackage{stfloats}
\usepackage{multirow}
\usepackage{tabularx}
\newcolumntype{Y}{>{\centering\arraybackslash}X}
\usepackage{fancyvrb}
\usepackage{arydshln}
\usepackage{makecell}
\usepackage{dsfont}
\usepackage{booktabs}
\usepackage{wrapfig}
\usepackage{enumitem}

\definecolor{rightgreen}{RGB}{0,154,24}

\usepackage[frozencache,cachedir=.]{minted} %

\newcommand{\overbar}[1]{\mkern 1.5mu\overline{\mkern-1.5mu#1\mkern-1.5mu}\mkern 1.5mu}

\usepackage[breaklinks=true,colorlinks,citecolor=black,bookmarks=false,urlcolor=blue]{hyperref}
\hypersetup{
	pdfinfo={
		Title={PixMix: Dreamlike Pictures Comprehensively Improve Safety Measures},
		Author={Dan Hendrycks and Andy Zou and Mantas Mazeika and Leonard Tang and Bo Li and Dawn Song and Jacob Steinhardt},
		Subject={Deep Learning, ML Safety, Robustness, Uncertainty},
		Keywords={robustness, uncertainty, out of distribution, ood detection, calibration, distribution shift, ml safety}
	}
}

\newcommand{\reviewer}[3]{
	\expandafter\newcommand\csname #1\endcsname[1]{
		\textcolor{#3}{[#2: ##1]}
	}
}
\reviewer{someone}{Someone}{red}
\definecolor{neonpurple}{rgb}{0.3,0,1}
\reviewer{dan}{Dan}{neonpurple}
\reviewer{js}{Jacob}{blue}
\reviewer{nicholas}{Nicholas}{red}
\reviewer{tom}{Tom}{green}

\usepackage[capitalize]{cleveref}
\crefname{section}{Sec.}{Secs.}
\Crefname{section}{Section}{Sections}
\Crefname{table}{Table}{Tables}
\crefname{table}{Tab.}{Tabs.}

\def\cvprPaperID{11590} %
\def\confName{CVPR}
\def\confYear{2022}

\makeatletter
\newcommand{\printfnsymbol}[1]{%
  \textsuperscript{\@fnsymbol{#1}}%
}
\makeatother

\begin{document}

\title{\textsc{PixMix}: Dreamlike Pictures\\Comprehensively Improve Safety Measures}

\author{Dan Hendrycks\thanks{Equal Contribution.}\\
UC Berkeley
\and
Andy Zou\printfnsymbol{1}\\
UC Berkeley
\and
Mantas Mazeika\\
UIUC
\and
Leonard Tang\\
Harvard University
\and
Bo Li\\
UIUC
\and
Dawn Song\\
UC Berkeley
\and
Jacob Steinhardt\\
UC Berkeley
}
\maketitle

\begin{abstract}
In real-world applications of machine learning, reliable and safe systems must consider measures of performance beyond standard test set accuracy. These other goals include out-of-distribution (OOD) robustness, prediction consistency, resilience to adversaries, calibrated uncertainty estimates, and the ability to detect anomalous inputs. However, improving performance towards these goals is often a balancing act that today's methods cannot achieve without sacrificing performance on other safety axes. For instance, adversarial training improves adversarial robustness but sharply degrades other classifier performance metrics. Similarly, strong data augmentation and regularization techniques often improve OOD robustness but harm anomaly detection, raising the question of whether a Pareto improvement on all existing safety measures is possible. To meet this challenge, we design a new data augmentation strategy utilizing the natural structural complexity of pictures such as fractals, which outperforms numerous baselines, is near Pareto-optimal, and roundly improves safety measures.
\end{abstract}

\input{sections/1-intro}

\input{sections/2-related_work}

\input{sections/3-method}
\input{sections/4-experiments}
\input{sections/5-conclusion}

\newpage
{\small
\bibliographystyle{ieee_fullname}
\bibliography{main}
}

\newpage
\appendix
\input{sections/6-appendix}

\end{document}

%% file: sections/1-intro.tex
\section{Introduction}

A central challenge in machine learning is building models that are reliable and safe in the real world. In addition to performing well on the training distribution, deployed models should be robust to distribution shifts, consistent in their predictions, resilient to adversaries, calibrated in their uncertainty estimates, and capable of identifying anomalous inputs. Numerous prior works have tackled each of these problems separately \cite{madry, hendrycks2019robustness, guo2017calibration, emmott2015meta}, but they can also be grouped together as various aspects of ML Safety \cite{hendrycks2021unsolved}. Consequently, the properties listed above can be thought of as safety measures.

\begin{figure}[t]
    \centering
    \hspace{-15pt}
    \includegraphics[width=0.5\textwidth]{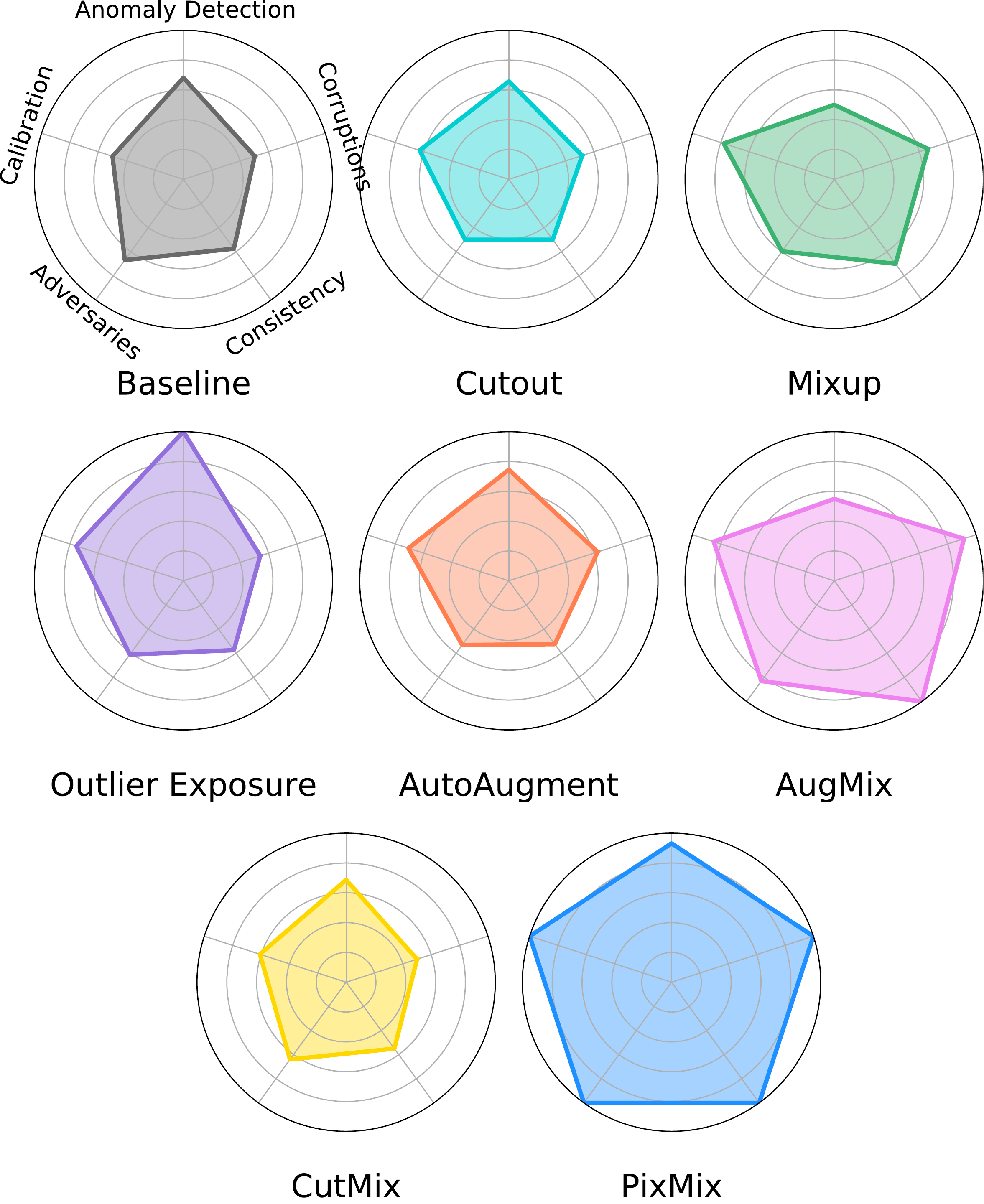}
    \caption{Normalized performance of different methods on five different model safety measures. \textsc{PixMix} is the only method that significantly outperforms the baseline in all five safety measures.}
    \label{fig:radar}
    \vspace{-10pt}
\end{figure}

Ideally, models deployed in real-world settings would perform well on multiple safety measures. Unfortunately, prior work has shown that optimizing for some desirable properties often comes at the cost of others. For example, adversarial training only improves adversarial robustness and degrades classification performance \cite{tsipras2018robustness}. Similarly, inducing consistent predictions on out-of-distribution (OOD) inputs seems to be at odds with better detecting these inputs, an intuition supported by recent work \cite{empiricalpaper} which finds that existing help with some safety metrics but harm others. %
This raises the question of whether improving all safety measures is possible with a single model.

\begin{table*}[t]
\vspace{-10pt}
\begin{center}
{
\setlength\tabcolsep{2.5pt}
\setlength\extrarowheight{1pt}
\renewcommand{\arraystretch}{2}
\begin{tabular}{l | c c c c c}
\multicolumn{1}{c}{} & \multicolumn{5}{c}{} \\%\multicolumn{1}{c}{}\\
\makecell{Method} & \makecell{Baseline \\ \includegraphics[width=0.15\textwidth]{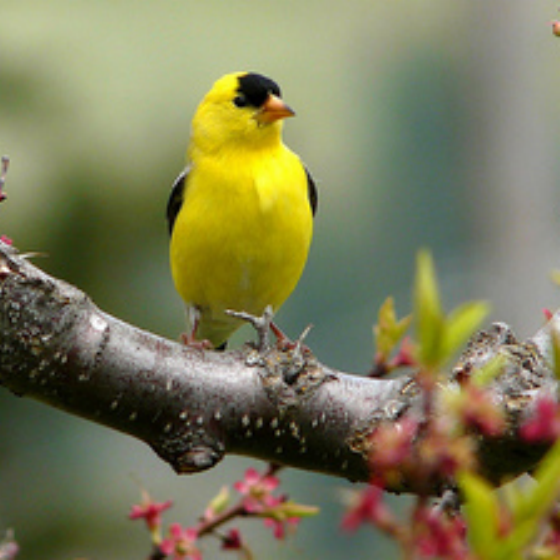}} & \makecell{Cutout \\ \includegraphics[width=0.15\textwidth]{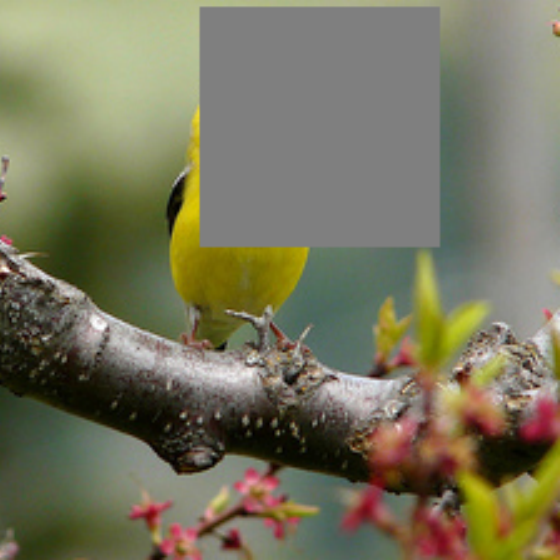}}	& \makecell{Mixup \\ \includegraphics[width=0.15\textwidth]{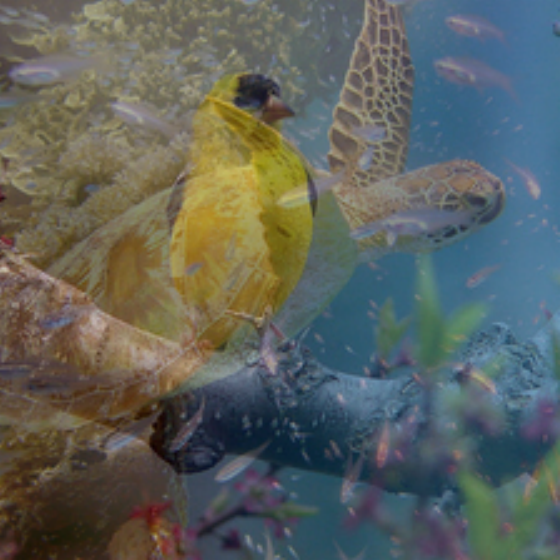}} & \makecell{CutMix \\ \includegraphics[width=0.15\textwidth]{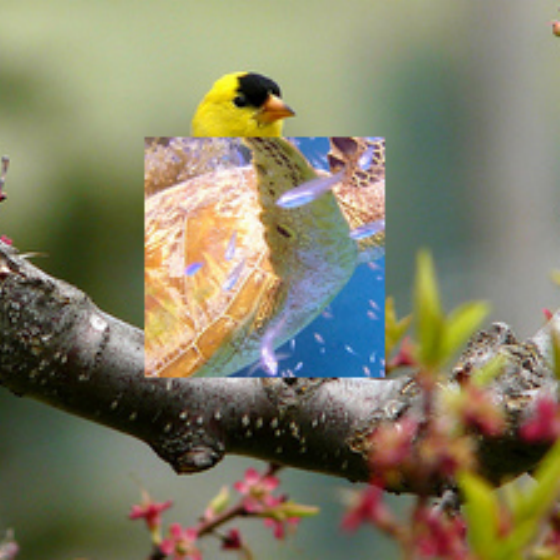}} & \makecell{\textsc{PixMix} \\ \includegraphics[width=0.15\textwidth]{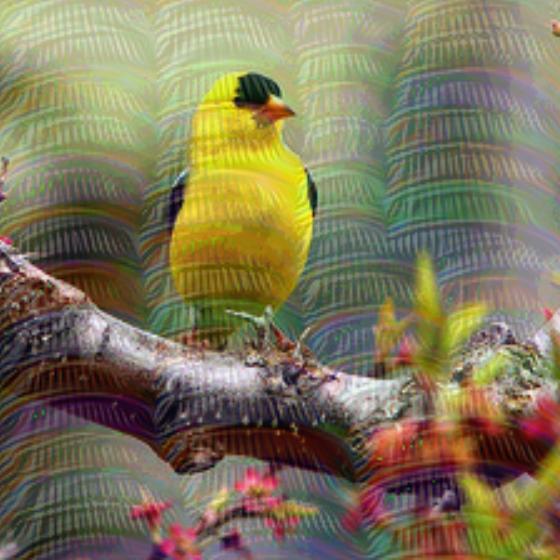}}\\
\hline
\makecell{Corruptions \\ \textcolor{gray}{mCE ($\downarrow$)}} & \makecell{50.0 \\ $+0.0$} & \makecell{51.5 \\ \color{red}{$+1.5$}} & \makecell{48.0 \\ \color{rightgreen}{$-2.0$}} & \makecell{51.5 \\ \color{red}{$+1.5$}} & \makecell{\textbf{30.5} \\ \color{rightgreen}{$\mathbf{-19.5}$}} \vspace{2pt}
\\
\hline
\makecell{Adversaries \\ \textcolor{gray}{Error ($\downarrow$)} } & \makecell{96.5 \\ $+0.0$} & \makecell{98.5 \\ \color{red}{$+1.0$}} & \makecell{97.4 \\ \color{red}{$+0.9$}} & \makecell{97.0 \\ \color{red}{$+0.5$}} & \makecell{\textbf{92.9} \\ \color{rightgreen}{$\mathbf{-3.9}$}} \vspace{2pt}
\\
\hline
\makecell{Consistency \\ \textcolor{gray}{mFR ($\downarrow$)} } & \makecell{10.7 \\ $+0.0$} & \makecell{11.9 \\ \color{red}{$+1.2$}} & \makecell{9.5 \\ \color{rightgreen}{$-1.2$}} & \makecell{12.0 \\ \color{red}{$+1.3$}} & \makecell{\textbf{5.7} \\ \color{rightgreen}{$\mathbf{-5.0}$}} \vspace{2pt}
\\
\hline
\makecell{Calibration \\ \textcolor{gray}{RMS Error ($\downarrow$)} } & \makecell{31.2 \\ $+0.0$} & \makecell{31.1 \\ \color{rightgreen}{$-0.1$}} & \makecell{13.0 \\ \color{rightgreen}{$-18.1$}} & \makecell{29.3 \\ \color{rightgreen}{$-1.8$}} & \makecell{\textbf{8.1} \\ \color{rightgreen}{$\mathbf{-23.0}$}} \vspace{2pt}
\\
\hline
\makecell{Anomaly Detection \\ \textcolor{gray}{AUROC ($\uparrow$)} } & \makecell{77.7 \\ $+0.0$} & \makecell{74.3 \\ \color{red}{$-3.4$}} & \makecell{71.7 \\ \color{red}{$-6.0$}} & \makecell{74.4 \\ \color{red}{$-3.3$}} & \makecell{$\mathbf{89.3}$ \\ \color{rightgreen}{$\mathbf{+11.6}$}} \vspace{2pt}
\\
\Xhline{2\arrayrulewidth}
\end{tabular}}
\caption{\textsc{PixMix} comprehensively improves safety measures, providing significant improvements over state-of-the-art baselines. We observe that previous augmentation methods introduce few additional sources of structural complexity. By contrast, \textsc{PixMix} incorporates fractals and feature visualizations into the training process, actively exposing models to new sources of structural complexity. We find that \textsc{PixMix} is able to improve both robustness and uncertainty estimation and is the first method to substantially improve all existing safety measures over the baseline.}
\vspace{-10pt}
\label{tab:big_results}
\end{center}
\end{table*}

While previous augmentation methods create images that are different (e.g., translations) or more entropic (e.g., additive Gaussian noise), we argue that an important underexplored axis is creating images that are more complex.
As opposed to entropy or descriptive difficulty, which is maximized by pure noise distributions, structural complexity is often described in terms of the degree of organization \cite{lloyd2001measures}. A classic example of structurally complex objects is fractals, which have recently proven useful for pretraining image classifiers \cite{kataoka2020pre, Nakashima2021CanVT}. Thus, an interesting question is whether sources of structural complexity can be leveraged to improve safety through data augmentation techniques.

We show that Pareto improvements are possible with \textsc{PixMix}, a simple and effective data processing method that leverages pictures with complex structures and substantially improves all existing safety measures. \textsc{PixMix} consists of a new data processing pipeline that incorporates structurally complex ``dreamlike'' images. These dreamlike images include fractals and feature visualizations. We find that feature visualizations are a suitable source of complexity, thereby demonstrating that they have uses beyond interpretability. In extensive experiments, we find that \textsc{PixMix} provides substantial gains on a broad range of existing safety measures, outperforming numerous previous methods. Code is available at \href{https://github.com/andyzoujm/pixmix}{\texttt{github.com/andyzoujm/pixmix}}.

%% file: sections/2-related_work.tex
\begin{figure*}[t]
    \centering
    \vspace{-20pt}
    \includegraphics[width=\textwidth]{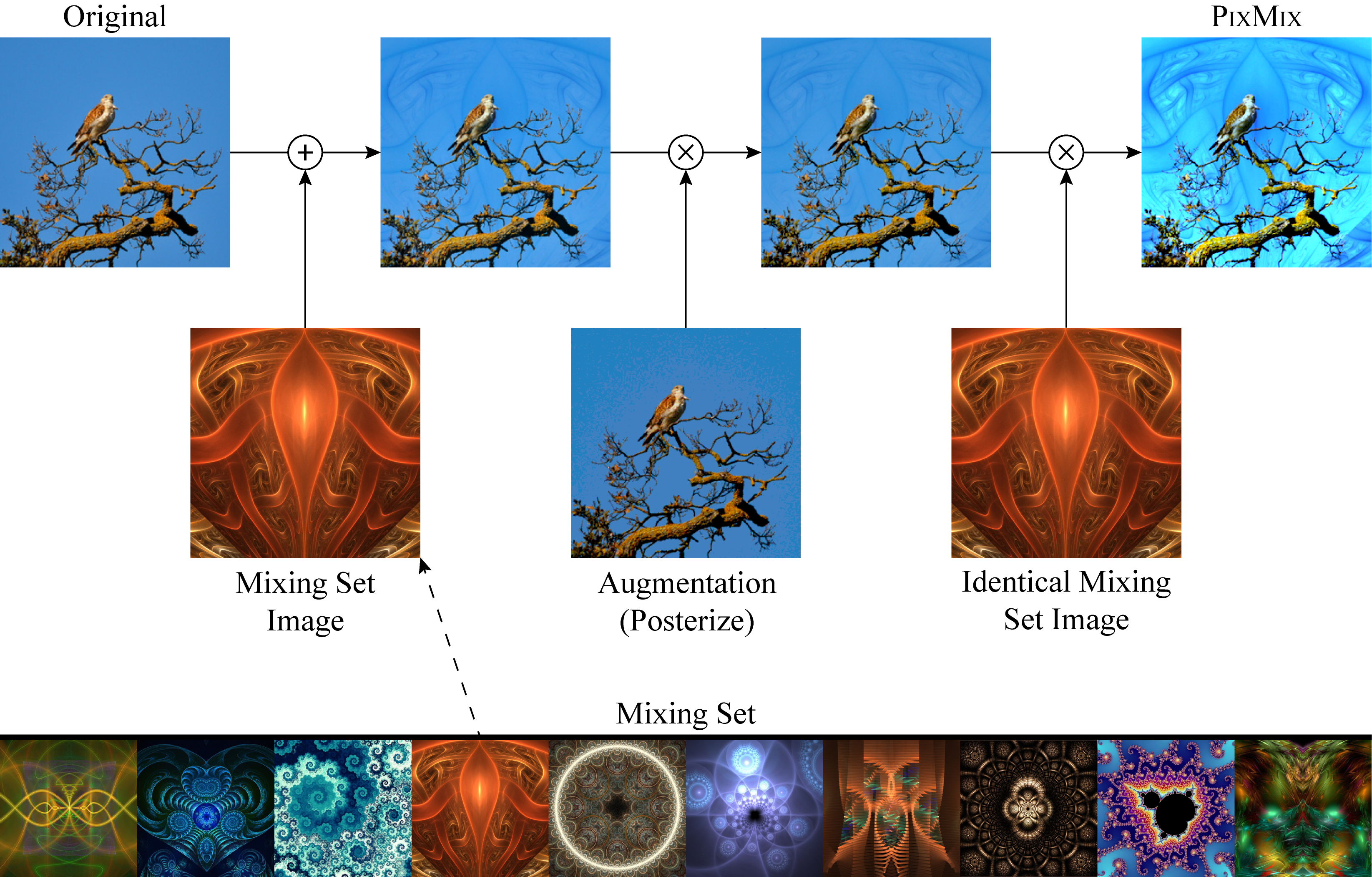}
    \caption{Top: An instance of a \textsc{PixMix} augmentation being applied to a bird image. The original clean image is mixed with augmented versions of itself and an image such as a fractal. Bottom: Sample images from the \textsc{PixMix} mixing set. We select fractals and feature visualizations from manually curated online sources. In ablations, we find that these new sources of visual structure for augmentations outperform numerous synthetic image distributions explored in prior work \cite{baradad2021learning}.}
    \label{fig:pixmix_pipeline}
\end{figure*}

\section{Related Work}

\noindent\textbf{Robustness.}\quad
Out-of-distribution robustness considers how to make ML models resistant to various forms of data shift at test time. Geirhos et al., 2019 \cite{geirhos2019} uncover a texture bias in convolutional networks and show that training on diverse stylized images can improve robustness at test-time. The ImageNet-C(orruptions) benchmark \cite{hendrycks2019robustness} consists of diverse image corruptions known to track robustness on some real world data shifts \cite{hendrycks2021many}. ImageNet-C is used to test models that are trained on ImageNet \cite{imagenet} and is used as a held-out, more difficult test set. They also introduce ImageNet-P(erturbations) for measuring prediction consistency under various non-adversarial input perturbations. Others have introduced additional corruptions for evaluation called ImageNet-$\overline{\text{C}}$ \cite{mintun2021interaction}. The ImageNet-R(enditions) benchmark measures performance degradation under various renditions of objects including paintings, cartoons, graffiti, embroidery, origami, sculptures, toys, and more \cite{hendrycks2021many}. In the similar setting of domain adaptation, Bashkirova et al., 2021 \cite{bashkirova2021visda} consider evaluating test-time robustness of models and even anomaly detection \cite{emmott2015meta,Liang2018EnhancingTR,ruff2021unifying}. Yin et al., 2019 \cite{Yin2019AFP} show that adversarial training can substantially reduce robustness on some corruptions and argue that part of model fragility is explained by overreliance on spurious cues \cite{Sagawa2019DistributionallyRN,Koh2021WILDSAB}.

\noindent\textbf{Calibration.}\quad
Calibrated prediction confidences are valuable for classification models in real-world settings. Several works have investigated evaluating and improving the calibration of deep neural networks \cite{nguyen2015posterior,guo2017calibration} through the use of validation sets. Others have shown that calibration can be improved without a validation set through methods such as ensembling \cite{lakshminarayanan2017simple} and pre-training \cite{hendrycks2019pretrain}. Ovadia et al. \cite{ovadia2019can} find that models are markedly less calibrated under distribution shift.

\noindent\textbf{Anomaly Detection.}\quad
Since models should ideally know what they do not know, they will need to identify when an example is anomalous. Anomaly detection seeks to estimate whether an input is out-of-distribution (OOD) with respect to a given training set. Hendrycks et al., 2017 \cite{hendrycks17baseline} propose a simple baseline for detecting classifier errors and OOD inputs. Devries et al., 2018 \cite{Devries2018LearningCF} propose training classifiers with an additional confidence branch for detecting OOD inputs. Lee et al., 2018 \cite{kimin} propose improving representations used for detectors with near-distribution images generated by GANs. Lee et al., 2018 \cite{mahal} also propose the Mahalanobis detector.
Outlier Exposure \cite{hendrycks2018deep} fine-tunes classifiers with diverse, natural anomalies, and since it is the state-of-the-art for OOD detection, we test this method in our paper. %

\noindent\textbf{Data Augmentation.}\quad
Simulated and augmented inputs can help make ML systems more robust, and this approach is used in real-world applications such as autonomous driving \cite{teslaaiday,waymo}. For state-of-the-art models, data augmentation can improve clean accuracy comparably to a $10\times$ increase in model size \cite{steiner2021train}. Further, data augmentation can improve out-of-distribution robustness comparably to a $1,\!000\times$ increase in labeled data \cite{hendrycks2021many}. Various augmentation techniques for image data have been proposed, including Cutout \cite{Devries2017ImprovedRO, Zhong2017RandomED}, Mixup \cite{Zhang2017mixupBE, tokozume2018between}, CutMix \cite{Yun2019CutMixRS, takahashi2019data}, and AutoAugment \cite{Cubuk2018AutoAugmentLA,Yin2019AFP}. Lopes et al., 2019 \cite{Lopes2019ImprovingRW} find that inserting random noise patches into training images improves robustness. AugMix is a data augmentation technique that specifically improves OOD generalization \cite{hendrycks2019augmix}. Chun et al. \cite{empiricalpaper} evaluates some of these techniques on CIFAR-10-C, a variant of ImageNet-C for the CIFAR-10 dataset \cite{hendrycks2019robustness}. They find that these data augmentation techniques can improve OOD generalization at the cost of weaker OOD detection.

\noindent\textbf{Analyzing Safety Goals Simultaneously.}\quad
Recent works study how a given method influences safety goals \cite{hendrycks2021unsolved} simultaneously. Prior work has shown that Mixup, CutMix, Cutout, ShakeDrop, adversarial training, Gaussian noise augmentation, and more have mixed effects on various safety metrics 
\cite{empiricalpaper}. Others have shown that different pretraining methods can improve some safety metrics and hardly affect others, but the pretraining method must be modified per task \cite{hendrycks2019pretrain}. Self-supervised learning methods can also be repurposed to help with some safety goals, all while not affecting others, but to realize the benefit, each task requires different self-supervised learning models \cite{hendrycks2019using}. Thus, creating a single method for improving performance across multiple safety metrics is an important next step.

\begin{figure*}[t]
    \vspace{-20pt}
    \begin{minted}[escapeinside=||,mathescape=true]{Python}
    def pixmix(|$x_\text{orig}$|, |$x_\text{mixing\_pic}$|, k=4, beta=3):
        |$x_\text{pixmix}$| = random.choice([|\color{blue}{augment}|(|$x_\text{orig}$|), |$x_\text{orig}$|])
    
        for i in range(random.choice([0,1,...,k])): # random count of mixing rounds
        
            # mixing_pic is from the mixing set (e.g., fractal, natural image, etc.)
            mix_image = random.choice([|\color{blue}{augment}|(|$x_\text{orig}$|), |$x_\text{mixing\_pic}$|])
            mix_op = random.choice([additive, multiplicative])
            
            |$x_\text{pixmix}$| = mix_op(|$x_\text{pixmix}$|, mix_image, beta)
        
        return |$x_\text{pixmix}$|
        
    def augment(x):
        aug_op = random.choice([rotate, solarize, ..., posterize])
        return aug_op(x)
    \end{minted}
    \caption{Simplified code for \textsc{PixMix}, our proposed data augmentation method. Initial images are mixed with a randomly selected image from our mixing set or augmentations of the clean image. The mixing operations are selected at random, and the mixing set includes fractals and feature visualization pictures. \textsc{PixMix} integrates new complex structures into the training process by leveraging fractals and feature visualizations, resulting in improved classifier robustness and uncertainty estimation across numerous safety measures.\looseness=-1
    }\label{fig:pseudocode}
    \vspace{-10pt}
\end{figure*}

\noindent\textbf{Training on Complex Synthetic Images.}\quad
Kataoka et al., 2020 \cite{kataoka2020pre} introduce FractalDB, a dataset of black-and-white fractals, and they show that pretraining on these algorithmically generated fractal images can yield better downstream performance than pretraining on many manually annotated natural datasets. Nakashima et al. \cite{Nakashima2021CanVT} show that models trained on a large variant of FractalDB can match ImageNet-1K pretraining on downstream tasks. Baradad et al., 2021 \cite{baradad2021learning} find that, for self-supervised learning, other synthetic datasets may be more effective than FractalDB, and they find that structural complexity and diversity are key properties for good downstream transfer. We depart from this recent line of work and ask whether structurally complex images can be repurposed for data augmentation instead of training from scratch. While data augmentation techniques such as those that add Gaussian noise increase input entropy, such noise has maximal \emph{descriptive} complexity but introduce little \emph{structural} complexity \cite{lloyd2001measures}. Since a popular definition of structural complexity is the fractal dimension \cite{lloyd2001measures}, we turn to fractals and other structurally complex images for data augmentation.

%% file: sections/3-method.tex
\section{Approach}
\noindent We propose \textsc{PixMix}, a simple and effective data augmentation technique that improves many ML Safety \cite{hendrycks2021unsolved} measures simultaneously, in addition to accuracy. \textsc{PixMix} is comprised of two main components: a set of structurally complex pictures (``Pix'') and a pipeline for augmenting clean training pictures (``Mix''). At a high level, \textsc{PixMix} integrates diverse patterns from fractals and feature visualizations into the training set. As fractals and feature visualizations do not belong to any particular class, we train networks to classify augmented images as the original class, as in standard data augmentation. %

\subsection{Picture Sources (\textsc{Pix})}
While \textsc{PixMix} can utilize arbitrary datasets of pictures, we discover that fractals and feature visualizations are especially useful pictures with complex structures. Collectively we refer to these two picture sources as ``dreamlike pictures.'' We analyze \textsc{PixMix} using other picture sources in the Appendix.

These pictures have ``non-accidental'' properties that humans may use, namely ``structural properties of contours (orientation, length, curvature) and contour junctions (types and angles) from line drawings of natural scenes'' \cite{Walther2014NonaccidentalPU}. Fractals possess some of these structural properties, and they are highly non-accidental and unlikely to arise from maximum entropy, unstructured random noise processes.

\noindent\textbf{Fractals.}\quad
Fractals can be generated in several ways, with one of the most common being iterated function systems. Rather than generate our own diverse fractals, which is a substantial research endeavor \cite{kataoka2020pre}, we download $14,\!230$ fractals from manually curated collections on DeviantArt. The resulting fractals are visually diverse, which can be seen in the bottom portion of \Cref{fig:pixmix_pipeline}.

\noindent\textbf{Feature Visualization.}\quad
Feature visualizations that maximize the response of neurons create archetypal images for neurons and often have high complexity \cite{45507, olah2017feature}. Thus, we include feature visualizations in our mixing set. We collect $4,\!700$ feature visualizations from the initial layers of several convolutional architectures using OpenAI Microscope. While feature visualizations have been primarily used for understanding network representations, we connect this line of interpretability work to improve performance on safety measures.

\subsection{Mixing Pipeline (\textsc{Mix})}
\noindent The pipeline for augmenting clean training images is described in Figure \ref{fig:pseudocode}. An instance of our mixing pipeline is shown in the top half of \Cref{fig:pixmix_pipeline}. First, a clean image has a $50\%$ chance of having a randomly selected standard augmentation applied. Next, we augment the image a random number of times with a maximum of $k$ times. Each augmentation is carried out by either additively or multiplicatively mixing the current image with a freshly augmented clean image or an image from the mixing set. Multiplicative mixing is performed similarly to the geometric mean. For both additive and multiplicative mixing, we use coefficients that are not convex combinations but rather conic combinations. Thus, additive and multiplicative mixing are performed with exponents and weights sampled from a Beta distribution independently. %

\begin{figure*}[t]
    \centering
    \vspace{-20pt}
    \includegraphics[width=\textwidth]{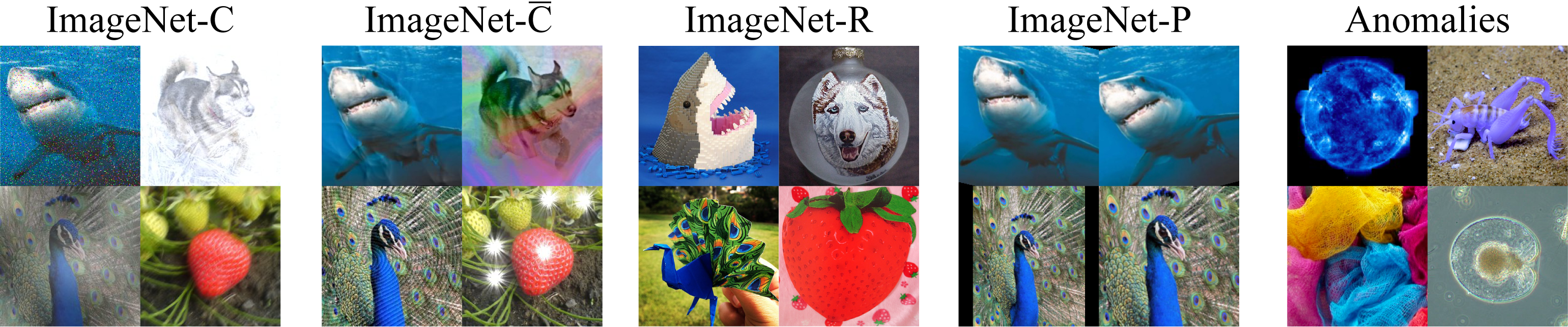}
    \caption{We comprehensively evaluate models across safety tasks, including corruption robustness (ImageNet-C, ImageNet-$\overline{\text{C}}$), rendition robustness (ImageNet-R), prediction consistency (ImageNet-P), confidence calibration, and anomaly detection. ImageNet-C \cite{hendrycks2019robustness} contains $15$ common corruptions, including fog, snow, and motion blur. ImageNet-$\overline{\text{C}}$ \cite{mintun2021interaction} contains additional corruptions. ImageNet-R \cite{hendrycks2021many} contains renditions of object categories and measures robustness to shape abstractions. ImageNet-P \cite{hendrycks2019robustness} contains sequences of gradual perturbations to images, across which predictions should be consistent. Anomalies are semantically distinct from the training classes. Existing work focuses on learning representations that improve performance on one or two metrics, often to the detriment of others. Developing models that perform well across multiple safety metrics is an important next step.
    }
    \label{fig:datasets}
    \vspace{-10pt}
\end{figure*}

%% file: sections/4-experiments.tex
\section{Experiments}

\noindent\textbf{Datasets.}\quad
We evaluate \textsc{PixMix} on extensions of CIFAR-10, CIFAR-100, and ImageNet-1K (henceforth referred to as ImageNet) for various safety tasks. So as not to ignore performance on the original tasks, we also evaluate on the standard versions of these datasets. %
ImageNet consists of $1.28$ million color images. As is common practice, we downsample ImageNet images to $224 \times 224$ resolution in all experiments. ImageNet consists of $1,\!000$ classes from WordNet noun synsets, covering a wide variety of objects, including fine-grained distinctions. We use the validation set for evaluating clean accuracy, which contains $50,\!000$ images.\looseness=-1

To measure corruption robustness, we use the CIFAR-10-C, CIFAR-100-C, and ImageNet-C datasets \cite{hendrycks2019robustness}. Each dataset consists of $15$ diverse corruptions applied to each image in the original test set. The corruptions can be grouped into blur, weather, and digital corruptions. Each corruption appears at five levels of severity. We also evaluate on the similar CIFAR-10-$\overbar{\text{C}}$ and ImageNet-$\overbar{\text{C}}$ datasets, which use a different set of corruptions \cite{mintun2021interaction}. To measure robustness to different renditions of object categories, we use the ImageNet-R dataset \cite{hendrycks2021many}. These datasets enable evaluating the out-of-distribution generalization of classifiers trained on clean data and non-overlapping augmentations.\looseness=-1

To measure consistency of predictions, we use the CIFAR-10-P, CIFAR-100-P, and ImageNet-P datasets. Each dataset consists of $10$ gradual shifts that images can undergo, such as zoom, translation, and brightness variation. Unlike other datasets we evaluate on, each example in these datasets is a video, and the objective is to have robust predictions that do not change across per-frame perturbations. These datasets enable measuring the stability, volatility, or ``jaggedness'' of network predictions in the face of minor perturbations. Examples from these datasets are in \Cref{fig:datasets}.

\noindent\textbf{Methods.}\quad
We compare \textsc{PixMix} to various state-of-the-art data augmentation methods. \textit{Baseline} denotes standard data augmentation; for ImageNet, we use the a random resized crop and random horizontal flipping, while on CIFAR-10 and CIFAR-100, we use random cropping with zero padding followed by random horizontal flips. \textit{Cutout} aims to improve representations by randomly masking out image patches, using patch side lengths that are half the side length of the original image. \textit{Mixup} regularizes networks to behave linearly between training examples by training on pixel-wise linear interpolations between input images and labels. \textit{CutMix} combines the techniques of Cutout and Mixup by replacing image patches with patches from other images in the training set. The labels of the resulting images are combined in proportion to the pixels taken by each source image. \textit{Auto Augment} searches for compositions of augmentations that maximize accuracy on a validation set. \textit{AugMix} uses a ResNeXt-like pipeline to combine randomly augmented images.
Compared to AugMix, which requires up to $9$ augmentations per image and can be slow to run, \textsc{PixMix} requires substantially fewer augmentations; we find an average of $2$ augmentations is sufficient. For fairness, we follow \cite{mintun2021interaction} and train AugMix without the Jensen-Shannon Divergence consistency loss, which requires at least thrice the memory per batch. \textit{Outlier Exposure} trains networks to be uncertain on a training dataset of outliers, and these outliers are distinct from the out-of-distribution test sets that we use during evaluation. For ImageNet experiments, we compare to several additional methods. \textit{SIN} trains networks on a mixture of clean images and images rendered using neural style transfer \cite{geirhos2019}. We opt for simple techniques that are widely used and do not evaluate all possible techniques from each of the areas we consider. More methods are evaluated in the Appendix.

\begin{table*}[ht]
\vspace{-10pt}
\setlength\tabcolsep{8pt}
\small
\centering
\begin{tabular}{ll | cccccccc } 
  & & Baseline & Cutout & Mixup & CutMix & \makecell{Auto\\Augment} & AugMix & \makecell{Outlier\\ Exposure} & \textsc{PixMix}
\\
\hline
\parbox[t]{3mm}{\multirow{5}{*}{\rotatebox{90}{CIFAR-10}}}
& Corruptions    & 26.4 & 25.9 & 21.0 & 26.5 & 22.2 & 12.4 & 25.1 & \noindent\textbf{9.5} \\
& Consistency   & 3.4 & 3.7 & 2.9 & 3.5 & 3.6 & 1.7 & 3.4 & \noindent\textbf{1.7} \\
& Adversaries   & 91.3 & 96.0 & 93.3 & 92.1 & 95.1 & 86.8 & 92.9 & \noindent\textbf{82.1} \\
& Calibration    & 22.7 & 17.8 & 12.1 & 18.6 & 14.8 & 9.4 & 13.0 & \noindent\textbf{3.7} \\
& Anomaly Detection ($\uparrow$)  & 91.9 & 91.4 & 88.2 & 92.0 & 93.2 & 89.2 & \noindent\textbf{98.4} & 97.0 \\
\Xhline{3\arrayrulewidth}
\parbox[t]{3mm}{\multirow{5}{*}{\rotatebox{90}{CIFAR-100}}}
& Corruptions    & 50.0 & 51.5 & 48.0 & 51.5 & 47.0 & 35.4 & 51.5 & \noindent\textbf{30.5} \\
& Consistency    & 10.7 & 11.9 & 9.5 & 12.0 & 11.2 & 6.5 & 11.3 & \noindent\textbf{5.7} \\
& Adversaries    & 96.8 & 98.5 & 97.4 & 97.0 & 98.1 & 95.6 & 97.2 & \noindent\textbf{92.9} \\
& Calibration    & 31.2 & 31.1 & 13.0 & 29.3 & 24.9 & 18.8 & 15.2 & \noindent\textbf{8.1} \\
& Anomaly Detection ($\uparrow$)    & 77.7 & 74.3 & 71.7 & 74.4 & 80.4 & 84.9 & \noindent\textbf{90.3} & 89.3 \\
\Xhline{3\arrayrulewidth}
\end{tabular}
\vspace{1pt}
\caption{On CIFAR-10 and CIFAR-100, \textsc{PixMix} outperforms state-of-the-art techniques on five distinct safety metrics. Lower is better except for anomaly detection, and full results are in the Supplementary Material. On robustness tasks and confidence calibration, \textsc{PixMix} outperforms all prior methods by significant margins. On anomaly detection, \textsc{PixMix} nearly matches the performance of the state-of-the-art Outlier Exposure method without requiring a large, diverse dataset of known outliers.}\label{tab:cifar}
\vspace{-10pt}
\end{table*}

\subsection{Tasks and Metrics}
We compare \textsc{PixMix} to methods on five distinct ML Safety tasks. Individual methods are trained on clean versions of CIFAR-10, CIFAR-100, and ImageNet. Then, they are evaluated on each of the following tasks.

\noindent\textbf{Corruptions.}\quad
This task is to classify corrupted images from the CIFAR-10-C, CIFAR-100-C, and ImageNet-C datasets. The metric is the mean corruption error (mCE) across all fifteen corruptions and five severities for each corruption. Lower is better.

\noindent\textbf{Consistency.}\quad
This task is to consistently classify sequences of perturbed images from CIFAR-10-P, CIFAR-100-P, and ImageNet-P. The main metric is the mean flip rate (mFR), which corresponds to the probability that adjacent images in a temporal sequence have different predicted classes. This can be written as $\mathbb{P}_{x \sim \mathcal{S}}(f(x_j) \neq f(x_{j-1}))$, where $x_i$ is the $i^{\text{th}}$ image in a sequence. For non-temporal sequences such as increasing noise values in a sequence $\mathcal{S}$, the metric is modified to $\mathbb{P}_{x \sim \mathcal{S}}(f(x_j) \neq f(x_1))$. Lower is better.

\noindent\textbf{Adversaries.}\quad
This task is to classify images that have been adversarially perturbed by projected gradient descent \cite{madry}. For this task, we focus on untargeted perturbations on CIFAR-10 and CIFAR-100 with an $\ell_\infty$ budget of $2/255$ and $20$ steps of optimization. We do not display results of ImageNet models against adversaries in our tables, as for all tested methods the accuracy declines to zero with this budget. The metric is the classifier error rate. Lower is better.

\noindent\textbf{Calibration.}\quad
This task is to classify images with calibrated prediction probabilities, i.e. matching the empirical frequency of correctness. For example, if a weather forecast predicts that it will rain with $70\%$ probability on ten occasions, then we would like the model to be correct $7/10$ times. Formally, we want posteriors from a model $f$ to satisfy $\mathbb{P}\left(Y = \arg\max_i f(X)_i \mid \max_i f(X)_i = C \right) = C$, where $X,Y$ are random variables representing the data distribution. The metric is RMS calibration error \cite{hendrycks2019oe}, which is computed as $\sqrt{\mathbb{E}_C[(\mathbb{P}(Y=\hat{Y} | C = c) - c)^2] }$, where $C$ is the classifier's confidence that its prediction $\hat{Y}$ is correct. We use adaptive binning \cite{Nguyen2015PosteriorCA} to compute this metric. Lower is better.

\noindent\textbf{Anomaly Detection.}\quad
In this task we detect out-of-distribution \cite{hendrycks17baseline} or out-of-class images from various unseen distributions. The anomaly distributions are Gaussian, Rademacher, Blobs, Textures \cite{textures}, SVHN \cite{SVHN}, LSUN \cite{lsun}, Places69 \cite{zhou2017places}. We describe each in the Appendix and report average AUROC. An AUROC of $50\%$ is random chance and $100\%$ is perfect detection. Higher is better.

\subsection{Results on CIFAR-10/100 Tasks}

\noindent\textbf{Training Setup.}\quad
In the following CIFAR experiments, we train a 40-4 Wide ResNet \cite{wideresnet} with a drop rate of $0.3$ for $100$ epochs. All experiments use an initial learning rate of $0.1$ which decays following a cosine learning rate schedule \cite{sgdr}. For \textsc{PixMix} experiments, we use $k=4, \beta=3$. Hyperparameter robustness is discussed in the Appendix. Additionally, we use a weight decay of $0.0001$ for Mixup and $0.0005$ otherwise.

\begin{table*}[ht]
\setlength\tabcolsep{5pt}
\small
\centering
\begin{tabular}{l | cccccccccccc } 
 & \multicolumn{1}{c}{Accuracy} & \multicolumn{3}{c}{Robustness} & \multicolumn{2}{c}{Consistency} & \multicolumn{4}{c}{Calibration} & \multicolumn{2}{c}{Anomaly Detection} \\ \cmidrule(lr){2-2} \cmidrule(lr){3-5} \cmidrule(lr){6-7} \cmidrule(lr){8-11} \cmidrule(lr){12-13}
 & Clean & C & $\overline{\text{C}}$ & R & \multicolumn{2}{c}{ImageNet-P} & Clean & C & $\overline{\text{C}}$ & R & \multicolumn{2}{c}{Out-of-Class Datasets} \\
 & \textcolor{gray}{Error} & \textcolor{gray}{mCE} & \textcolor{gray}{Error} & \textcolor{gray}{Error} & \textcolor{gray}{mFR} & \textcolor{gray}{mT5D} &  \textcolor{gray}{RMS} & \textcolor{gray}{RMS} & \textcolor{gray}{RMS} & \textcolor{gray}{RMS} & \textcolor{gray}{AUROC ($\uparrow$)} & \textcolor{gray}{AUPR ($\uparrow$)}
\\
\hline
Baseline & 23.9&	78.2&	61.0& 63.8 &	58.0&	78.4&	5.6&	12.0&   20.7&	19.7&	79.7& 48.6 \\
Cutout & \underline{22.6} &	76.9&	60.2&	64.8 &  57.9&	75.2&	3.8&	11.1&	17.1&	14.6&	81.7& 49.6 \\
Mixup & 22.7&	72.7&	55.0&   62.3&	54.3&	73.2&	5.8&	7.3&	13.2&	44.6&	72.2& 51.3 \\
CutMix & 22.9&	77.8&	59.8&	66.5 & 60.3 &	76.6&	6.2&	9.1&	15.3&	43.5&	78.4& 47.9 \\
AutoAugment & \textbf{22.4}&	73.8&	58.0&	61.9 &  54.2&	72.0&	\textbf{3.6}&	8.0&	14.3&	12.6&	84.4& 58.2\\
AugMix & 22.8&	71.0&	56.5&   61.7&	52.7&	70.9&	4.5&	9.2&	15.0&	13.2&	84.2& 61.1 \\
SIN & 25.4&	70.9&	57.6&	\textbf{58.5}&   54.4&	71.8&	4.2&	6.5&	14.0&	16.2&	84.8& 62.3 \\
\textsc{PixMix} & \underline{22.6}&	\textbf{65.8}&	\textbf{44.3}&	\underline{60.1} & \textbf{51.1}&	\textbf{69.1}&	\textbf{3.6}&	\textbf{6.3}&	\textbf{5.8}&	\textbf{11.0}&	\textbf{85.7}& \textbf{64.1} \\
\Xhline{3\arrayrulewidth}
\end{tabular}
\caption{On ImageNet, \textsc{PixMix} improves over state-of-the-art methods on a broad range of safety metrics. Lower is better except for anomaly detection, and the full results are in the Supplementary Material. \textbf{Bold} is best, and \underline{underline} is second best. Across evaluation settings, \textsc{PixMix} is occasionally second-best, but it is usually first, making it near Pareto-optimal.}\label{tab:imagenet_results}
\vspace{-10pt}
\end{table*}

\noindent\textbf{Results.}\quad
In Table \ref{tab:big_results}, we see that \textsc{PixMix} improves over the standard baseline method on all safety measures. Moreover, all other methods decrease performance relative to the baseline for at least one metric, while \textsc{PixMix} is the first method to improve performance in all settings. Results for all other methods are in Table \ref{tab:cifar}. \textsc{PixMix} obtains better performance than all methods on Corruptions, Consistency, Adversaries, and Calibration. Notably, \textsc{PixMix} is far better than other methods for improving confidence calibration, reaching acceptably low calibration error on CIFAR-10. For corruption robustness, performance improvements on CIFAR-100 are especially large, with mCE on the Corruptions task dropping by $4.9\%$ compared to AugMix and $19.5\%$ compared to the baseline.

In addition to robustness and calibration, \textsc{PixMix} also greatly improves anomaly detection. \textsc{PixMix} nearly matches the anomaly detection performance of Outlier Exposure, the state-of-the-art anomaly detection method, without requiring large quantities of diverse, known outliers. This is surprising, as \textsc{PixMix} uses a standard cross-entropy loss, which makes the augmented images seem more in-distribution. Hence, one might expect unseen corruptions to be harder to distinguish as well, but in fact we observe the opposite---anomalies are easier to distinguish. Additional results and ablations are in the Appendix.

\begin{table*}[ht]
\setlength\tabcolsep{8pt}
\small
\centering
\begin{tabular}{ll | cccccc } 
& & \multicolumn{1}{c}{Accuracy} & \multicolumn{1}{c}{Corruptions} & \multicolumn{1}{c}{Consistency} & \multicolumn{1}{c}{Adversaries} & \multicolumn{1}{c}{Calibration} & \multicolumn{1}{c}{Anomaly} \\
\cmidrule(lr){3-3} \cmidrule(lr){4-4} \cmidrule(lr){5-5} \cmidrule(lr){6-6} \cmidrule(lr){7-7} \cmidrule(lr){8-8}
& & Clean & C & CIFAR-P & PGD & C & Detection \\
& \textsc{PixMix} Mixing Set & \textcolor{gray}{Error} & \textcolor{gray}{mCE} & \textcolor{gray}{mFR} & \textcolor{gray}{Error} & \textcolor{gray}{RMS} & \textcolor{gray}{AUROC ($\uparrow$)}
\\
\hline
\parbox[t]{1mm}{\multirow{5}{*}{\rotatebox{90}{Previous}}}
& Dead Leaves (Squares) \cite{baradad2021learning} &21.3&	36.2&	6.3&	94.1&	15.8&	81.8\\
& Spectrum + Color + WMM \cite{baradad2021learning} &20.7&	36.1&	6.6&	94.4&	15.9&	85.8\\
& StyleGAN (Oriented) \cite{baradad2021learning} &20.4&	37.3&	7.2&	97.0&	14.9&	83.7\\
& FractalDB \cite{kataoka2020pre} &\underline{20.3}&	33.9&	6.4&	98.2&	12.0&	82.5\\
& 300K Random Images \cite{hendrycks2019oe} &\textbf{19.6}&	34.5&	6.3&	94.7&	12.9&	86.2\\
\cdashline{1-8}
\parbox[t]{1mm}{\multirow{3}{*}{\rotatebox{90}{New}}}
& Fractals &\underline{20.3}&	32.3&	6.2&	95.5&	\underline{8.7}&	\underline{88.9}\\
& Feature Visualization (FVis) &21.5&	\textbf{30.3}&	\textbf{5.4}&	\textbf{91.5}&	9.9&	88.1\\
& Fractals + FVis &\underline{20.3}&	\underline{30.5}&	\underline{5.7}&	\underline{92.9}&	\textbf{8.1} &  \textbf{89.3}\\
\Xhline{3\arrayrulewidth}
\end{tabular}
\caption{Mixing set ablations showing that \textsc{PixMix} can use numerous mixing sets, including real images. Results are using CIFAR-100. \textbf{Bold} is best, and \underline{underline} is second best. We compare Fractals + FVis, the mixing set used as \textsc{PixMix}'s default mixing set, to other datasets from prior work. The 300K Random Images are real images scraped from online for Outlier Exposure. We discover the distinct utility of Fractals and FVis. By utilizing the 300K Random Images mixing set, \textsc{PixMix} can attain a $19.6\%$ error rate, though fractals can provide more robustness than these real images.\looseness=-1}\label{tab:dataset_ablation}
\end{table*}

\subsection{Results on ImageNet Tasks}

\noindent\textbf{Training Setup.}\quad
Since regularization methods may require a greater number of training epochs to converge, we fine-tune a pre-trained ResNet-50 for 90 epochs. For \textsc{PixMix} experiments, we use $k=4, \beta=4$. We use a batch size of $512$ and an initial learning rate of $0.01$ following a cosine decay schedule.

\noindent\textbf{Results.}\quad
We show ImageNet results in Table \ref{tab:imagenet_results}. Compared to the standard augmentations of the baseline, \textsc{PixMix} has higher performance on all safety measures. By contrast, other augmentation methods have lower performance than the baseline (cropping and flipping) on some metrics. Thus, \textsc{PixMix} is the first augmentation method with a Pareto improvement over the baseline on a broad range of safety measures.

On corruption robustness, \textsc{PixMix} outperforms state-of-the-art augmentation methods such as AugMix, improving mCE by $12.4\%$ over the baseline and $5.1\%$ over the mCE of the next-best method. On rendition robustness, \textsc{PixMix} outperforms all other methods save for SIN. Note that SIN is particularly well-suited to improving rendition robustness, as it trains on stylized ImageNet data. However, SIN incurs a $2\%$ loss to clean accuracy, while \textsc{PixMix} increases clean accuracy by $1.3\%$. Maintaining strong performance on clean images is an important property for methods to have, as practitioners may be unwilling to adopt methods that markedly reduce performance in ideal conditions.

\begin{figure}[]
    \centering
    \includegraphics[width=0.48\textwidth]{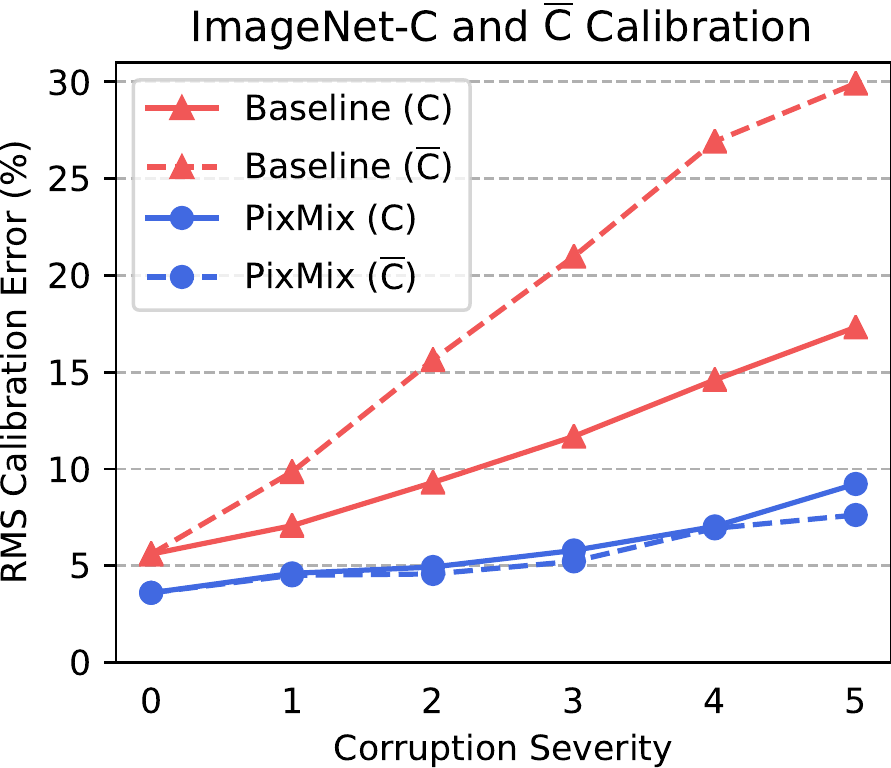}
    \caption{As corruption severity increases, \textsc{PixMix} calibration error increases much more slowly than the baseline calibration error, demonstrating that \textsc{PixMix} can improve uncertainty estimation under distribution shifts with unseen image corruptions.}
    \label{fig:calibration}
    \vspace{-15pt}
\end{figure}

On calibration tasks, \textsc{PixMix} outperforms all methods. As Ovadia et al. \cite{ovadia2019can} show, models are markedly less calibrated under distribution shift. We find that \textsc{PixMix} cuts calibration error in half on ImageNet-C compared to the baseline. On ImageNet-$\overbar{\text{C}}$, the improvement is even larger, with a $14.9\%$ reduction in absolute error. In \Cref{fig:calibration}, we visualize how calibration error on ImageNet-C and ImageNet-$\overbar{\text{C}}$ varies as the corruption severities increase. Compared to the baseline, \textsc{PixMix} calibration error increases much more slowly.
Further uncertainty estimation results are in the Appendix. For example, \textsc{PixMix} substantially improves anomaly detection performance with Places365 as the in-distribution set.

\subsection{Mixing Set Picture Source Ablations}
While we provide a high-quality source of structural complexity with \textsc{PixMix}, our mixing pipeline could be used with other mixing sets. %
In \Cref{tab:dataset_ablation}, we analyze the choice of mixing set on CIFAR-100 performance. We replace our Fractals and Feature Visualizations dataset (Fractals + FVis) with several synthetic datasets developed for unsupervised representation learning \cite{baradad2021learning, kataoka2020pre}. We also evaluate the 300K Random Images dataset of natural images used for Outlier Exposure on CIFAR-10 and CIFAR-100 \cite{hendrycks2019oe}.

Compared to alternative sources of visual structure, the Fractals + FVis mixing set yields substantially better results. This suggests that structural complexity in the mixing set is important. Indeed, the next-best method for reducing mCE on CIFAR-100-C is FractalDB, which consists of weakly curated black-and-white fractal images. By contrast, our Fractals dataset consists of color images of fractals that were manually designed and curated for being visually interesting. Furthermore, we find that removing either Fractals or FVis from the mixing set yields lower performance on safety metrics or lower performance on clean data, showing that both components of our mixing set are important. Similar ablations on ImageNet shown in \Cref{tab:imagenet_ablation} follow the same trend.

%% file: sections/5-conclusion.tex
\section{Conclusion}
\noindent We proposed \textsc{PixMix}, a simple and effective data augmentation technique for improving ML safety measures. Unlike previous data augmentation techniques, \textsc{PixMix} introduces new complexity into the training procedure by leveraging fractals and feature visualizations. We evaluated \textsc{PixMix} on numerous distinct ML Safety tasks: corruption robustness, rendition robustness, prediction consistency, adversarial robustness, confidence calibration, and anomaly detection. We found that \textsc{PixMix} was the first method to provide substantial improvements over the baseline on all existing safety metrics, and it obtained state-of-the-art performance in nearly all settings.

%% file: sections/6-appendix.tex
\newpage

\section{Additional Results}

\paragraph{Mixing Strategies.}
In \Cref{tab:cifar_appendix1}, we analyze different mixing strategies. The full \textsc{PixMix} mixing strategy is depicted in Figures 2 and 3 of the main paper. Mix Input only includes clean images in the mixing pipeline and does not use the mixing set at all. This severely harms performance on all safety metrics. Mix Aug only mixes with images from the mixing set. This reduces RMS calibration error but increases error on robustness tasks compared to \textsc{PixMix} Original. Finally, Iterative mixes with feature visualizations computed on the fly for the network being trained. This performs well on robustness tasks but has weaker calibration and anomaly detection. Additionally, computing feature visualizations at each iteration of training is substantially slower than precomputing them on fixed networks as we do in \textsc{PixMix}.

\paragraph{Full Results.}
In Tables \ref{tab:cifar_appendix3}, \ref{tab:cifar_appendix4}, and \ref{tab:imagenet_appendix}, we report full results for CIFAR-10, CIFAR-100, and ImageNet. The ImageNet results are copied from the main paper. For CIFAR, we evaluate on additional datasets, including CIFAR-10-$\overbar{\text{C}}$ and CIFAR-100-$\overbar{\text{C}}$, additional datasets of corrupted CIFAR images. We also report the mT5D metric on ImageNet-P. In all cases, \textsc{PixMix} provides the best overall performance.

\paragraph{Noise-Based Augmentations.}
Since noise-based augmentations sometimes nearly overlap with the test distribution and thereby may have an unfair advantage, we separately compare to several additional baselines on ImageNet that use noise-based data augmentations. \textit{ANT} trains networks on inputs with adversarially transformed noise applied \cite{rusak2020simple}. \textit{Speckle} trains on inputs with speckle noise added, which has been observed to improve robustness. \textit{EDSR} and \textit{Noise2Net} inject noise using image-to-image neural networks with noisy parameters \cite{hendrycks2021many}. \textit{Adversarial} trains networks with $\ell_\infty$ perturbations of magnitude $\varepsilon = 8/255$ \cite{madry2017towards}.

Results are in Tables \ref{tab:noise_methods}. %
We find that ANT and Speckle have strong performance on ImageNet-P overall, but this mostly comes from the Gaussian and shot noise categories. If we only consider prediction stability on non-noise categories, \textsc{PixMix} exhibits the least volatility in predictions out of all the methods considered.

\paragraph{Hyperparameter Sensitivity.}
In Table \ref{tab:cifar_appendix5}, we examine the hyperparameter sensitivity of \textsc{PixMix} on corruption robustness for CIFAR-100. We vary the $\beta$ and $k$ hyperparameters and find that performance is very stable across a range of hyperparameters.

\paragraph{Places365 Anomaly Detection.}
In Table \ref{tab:appendix_anomaly_detection}, we show anomaly detection performance with Places365 as the in-distribution data. For all methods, we use a ResNet-18 pre-trained on Places365. \textsc{PixMix} and Outlier Exposure (OE) are fine-tuned for $10$ epochs. We find that \textsc{PixMix} nearly matches the state-of-the-art OE detector despite being a general data augmentation technique that improves many other safety metrics.

\section{Outlier Datasets}
For anomaly detection, we use a suite of out-of-distribution datasets and average metrics across all OOD datasets in the main results. Gaussian noise is IID noise sampled from a normal distribution. Rademacher Noise is noise with each pixel sampled from $\{-1,1\}$ with equal probability. Blobs are algorithmically generated blobs. Textures are from the Describable Textures Dataset \cite{textures}. SVHN has images of numbers from houses. Places69 contains 69 scene categories and is disjoint from Places365.

\section{Broader Impacts}
As \textsc{PixMix} differentially improves safety metrics, it could have various beneficial effects. Improved robustness can result in more reliable machine learning systems deployed in safety-critical situations \cite{hendrycks2021unsolved}, such as self-driving cars. Anomaly detection enables better human oversight of machine learning systems and fallback policies in cases where systems encounter inputs they were not designed to handle. At the same time, anomaly detection could be misused as a surveillance tool, requiring careful consideration of individual use cases. Calibration enables more meaningful predictions that increase trust with end users. Additionally, compared to other methods for improving robustness, \textsc{PixMix} requires minimal modification of the training setup and a low computational overhead, resulting in lower costs to machine learning practitioners and the environment.

\begin{table*}[ht]
\setlength\tabcolsep{8pt}
\small
\centering
\begin{tabular}{l | cccccc }
 & \multicolumn{1}{c}{Accuracy} & \multicolumn{1}{c}{Corruptions} & \multicolumn{1}{c}{Consistency} & \multicolumn{1}{c}{Adversaries} & \multicolumn{1}{c}{Calibration} & \multicolumn{1}{c}{Anomaly} \\ \cmidrule(lr){2-2} \cmidrule(lr){3-3} \cmidrule(lr){4-4} \cmidrule(lr){5-5} \cmidrule(lr){6-6} \cmidrule(lr){7-7}
 & Clean & C & CIFAR-P & PGD & C & Detection \\
 & \textcolor{gray}{Error ($\downarrow$)} & \textcolor{gray}{mCE ($\downarrow$)} & \textcolor{gray}{mFR ($\downarrow$)} & \textcolor{gray}{Error ($\downarrow$)} & \textcolor{gray}{RMS ($\downarrow$)} & \textcolor{gray}{AUROC ($\uparrow$)}
\\
\hline
\textsc{PixMix} Original & 20.3&	30.5&	5.7&	92.9&	8.1&	89.3\\
Mix Input & 19.9&	34.1&	6.4&	96.7&	15.5&	86.5 \\
Mix Aug & 20.6&	31.1&	6.2&	94.2&	6.0&	89.7 \\
Iterative & 21.1&	31.4&	5.6&	90.6&	12.7&	86.7 \\
\Xhline{3\arrayrulewidth}
\end{tabular}
\caption{\textsc{PixMix} variations on CIFAR-100. Mix Input only mixes with augmented versions of the clean image. Mix Aug only mixes with images from the mixing set (i.e. fractals and feature visualizations). Iterative mixes with feature visualizations computed on the fly for the current network. Using the mixing set alone is more effective than augmented images alone, and combining them can further improve performance on several metrics.}\label{tab:cifar_appendix1}
\end{table*}

\begin{table*}[ht]
\setlength\tabcolsep{5pt}
\small
\centering
\begin{tabular}{l | ccccccccccc } 
 & \multicolumn{1}{c}{Accuracy} & \multicolumn{2}{c}{Corruptions} & \multicolumn{2}{c}{Consistency} & \multicolumn{1}{c}{Adversaries} & \multicolumn{3}{c}{Calibration} & \multicolumn{2}{c}{Anomaly} \\ \cmidrule(lr){2-2} \cmidrule(lr){3-4} \cmidrule(lr){5-6} \cmidrule(lr){7-7} \cmidrule(lr){8-10} \cmidrule(lr){11-12}
 & Clean & C & $\overline{\text{C}}$ & \multicolumn{2}{c}{CIFAR-P} & PGD & Clean & C & $\overline{\text{C}}$ & \multicolumn{2}{c}{Detection} \\
 & \textcolor{gray}{Error} & \textcolor{gray}{mCE} & \textcolor{gray}{mCE} & \textcolor{gray}{mFR} & \textcolor{gray}{mT5D} & \textcolor{gray}{Error} & \textcolor{gray}{RMS} & \textcolor{gray}{RMS} & \textcolor{gray}{RMS} & \textcolor{gray}{AUROC ($\uparrow$)} & \textcolor{gray}{AUPR ($\uparrow$)}
\\
\hline
CutMix   & 20.3&	51.5&	49.6&	12.0&	3.0&	97.0&	12.2&	29.3&	26.5&	74.4& 32.3 \\
\textsc{PixMix}   & 20.3&	\textbf{30.5}&	36.7&	\textbf{5.7}&	\textbf{1.6}&	\textbf{92.9}&	7.0&	8.1&	8.9&	89.3& \textbf{70.9} \\
\textsc{PixMix} + CutMix   & \textbf{19.9}&	30.9&	\textbf{35.5}&	5.8&	1.7&	93.1&	\textbf{4.4}&	\textbf{6.0}&	\textbf{5.9}&	\textbf{89.5}& 68.6 \\
\Xhline{3\arrayrulewidth}
\end{tabular}
\caption{Combining \textsc{PixMix} and CutMix on CIFAR-100. While \textsc{PixMix} is strong on its own, combination with other data augmentation techniques can further improve performance.}\label{tab:cifar_appendix3}
\end{table*}

\begin{table*}[ht]
\setlength\tabcolsep{5pt}
\small
\centering
\begin{tabular}{l | ccccccccccc } 
 & \multicolumn{1}{c}{Accuracy} & \multicolumn{2}{c}{Corruptions} & \multicolumn{2}{c}{Consistency} & \multicolumn{1}{c}{Adversaries} & \multicolumn{3}{c}{Calibration} & \multicolumn{2}{c}{Anomaly} \\ \cmidrule(lr){2-2} \cmidrule(lr){3-4} \cmidrule(lr){5-6} \cmidrule(lr){7-7} \cmidrule(lr){8-10} \cmidrule(lr){11-12}
 & Clean & C & $\overline{\text{C}}$ & \multicolumn{2}{c}{CIFAR-P} & PGD & Clean & C & $\overline{\text{C}}$ & \multicolumn{2}{c}{Detection} \\
 & \textcolor{gray}{Error} & \textcolor{gray}{mCE} & \textcolor{gray}{mCE} & \textcolor{gray}{mFR} & \textcolor{gray}{mT5D} & \textcolor{gray}{Error} & \textcolor{gray}{RMS} & \textcolor{gray}{RMS} & \textcolor{gray}{RMS} & \textcolor{gray}{AUROC ($\uparrow$)} & \textcolor{gray}{AUPR ($\uparrow$)}
\\
\hline
Baseline   & 21.3&	50.0&	52.0&	10.7&	2.7&	96.8&	14.6&	31.2&	30.9&	77.7& 35.4 \\
Cutout   & 19.9&	51.5&	50.2&	11.9&	2.7&	98.5&	11.4&	31.1&	29.4&	74.3& 31.3 \\
Mixup   & 21.1&	48.0&	49.8&	9.5&	3.0&	97.4&	10.5&	13.0&	12.9&	71.7& 31.9 \\
CutMix   & 20.3&	51.5&	49.6&	12.0&	3.0&	97.0&	12.2&	29.3&	26.5&	74.4& 32.3 \\
AutoAugment  & \textbf{19.6}&	47.0&	46.8&	11.2&	2.6&	98.1&	9.9&	24.9&	22.8&	80.4& 33.2  \\
AugMix   & 20.6&	35.4&	41.2&	6.5&	1.9&	95.6&	12.5&	18.8&	22.5&	84.9& 53.8 \\
OE  & 21.9&	50.3&	52.1&	11.3&	3.0&	97.0&	12.0&	13.8&	13.9&	\textbf{90.3}& 66.2 \\
\textsc{PixMix}   & 20.3&	\textbf{30.5}&	\textbf{36.7}&	\textbf{5.7}&	\textbf{1.6}&	\textbf{92.9}&	\textbf{7.0}&	\textbf{8.1}&	\textbf{8.9}&	89.3& \textbf{70.9} \\
\Xhline{3\arrayrulewidth}
\end{tabular}
\caption{Full results for CIFAR-100. mT5D is an additional metric used for gauging prediction consistency in ImageNet-P, which we adapt to CIFAR-100.
Note \textsc{PixMix} can achieve 19.6\% error rate if it uses 300K Random Images as the Mixing Set, so \textsc{PixMix} can achieve the same accuracy as AutoAugment yet also do better on safety metrics.
}\label{tab:cifar_appendix3}
\end{table*}

\begin{table*}[ht]
\setlength\tabcolsep{5pt}
\small
\centering
\begin{tabular}{l | ccccccccccc } 
 & \multicolumn{1}{c}{Accuracy} & \multicolumn{2}{c}{Corruptions} & \multicolumn{2}{c}{Consistency} & \multicolumn{1}{c}{Adversaries} & \multicolumn{3}{c}{Calibration} & \multicolumn{2}{c}{Anomaly} \\ \cmidrule(lr){2-2} \cmidrule(lr){3-4} \cmidrule(lr){5-6} \cmidrule(lr){7-7} \cmidrule(lr){8-10} \cmidrule(lr){11-12}
 & Clean & CIFAR-C & $\overline{\text{C}}$ & \multicolumn{2}{c}{CIFAR-P} & PGD & Clean & CIFAR-C & $\overline{\text{C}}$ & \multicolumn{2}{c}{Detection} \\
 & \textcolor{gray}{Error} & \textcolor{gray}{mCE} & \textcolor{gray}{mCE} & \textcolor{gray}{mFR} & \textcolor{gray}{mT5D} & \textcolor{gray}{Error} & \textcolor{gray}{RMS} & \textcolor{gray}{RMS} & \textcolor{gray}{RMS} & \textcolor{gray}{AUROC ($\uparrow$)} & \textcolor{gray}{AUPR ($\uparrow$)}
\\
\hline
Baseline & 4.4&	26.4&	26.4&	3.4&	1.7&	91.3&	6.4&	22.7&	22.4&	91.9& 70.9 \\
Cutout & \textbf{3.6}&	25.9&	24.5&	3.7&	1.7&	96.0&	3.3&	17.8&	17.5&	91.4& 63.6 \\
Mixup & 4.2&	21.0&	22.1&	2.9&	2.1&	93.3&	12.5&	12.1&	10.9&	88.2& 67.1 \\
CutMix & 4.0&	26.5&	25.4&	3.5&	2.1&	92.1&	5.0&	18.6&	17.8&	92.0& 65.5 \\
AutoAugment & 3.9&	22.2&	24.4&	3.6&	1.7&	95.1&	4.0&	14.8&	16.6&	93.2& 64.6 \\
AugMix & 4.3&	12.4&	16.4&	1.7&	1.2&	86.8&	5.1&	9.4&	12.6&	89.2& 61.5 \\
OE & 4.6&	25.1&	26.1&	3.4&	1.9&	92.9&	6.9&	13.0&	13.2&	\textbf{98.4}& \textbf{92.5} \\
\textsc{PixMix} & 4.2&	\textbf{9.5}&	\textbf{13.6}&	\textbf{1.7}&	\textbf{1.0}&	\textbf{82.1}&	\textbf{2.6}&	\textbf{3.7}&	\textbf{5.3}&	97.0& 88.4 \\
\Xhline{3\arrayrulewidth}
\end{tabular}
\caption{Full results for CIFAR-10. mT5D is an additional metric used for gauging prediction consistency in ImageNet-P, which we adapt to CIFAR-10.}\label{tab:cifar_appendix4}
\end{table*}

\begin{table*}[ht]
\setlength\tabcolsep{5pt}
\small
\centering
\begin{tabular}{l | cccccccccccc } 
 & \multicolumn{1}{c}{Accuracy} & \multicolumn{3}{c}{Robustness} & \multicolumn{2}{c}{Consistency} & \multicolumn{4}{c}{Calibration} & \multicolumn{2}{c}{Anomaly} \\ \cmidrule(lr){2-2} \cmidrule(lr){3-5} \cmidrule(lr){6-7} \cmidrule(lr){8-11} \cmidrule(lr){12-13}
 & Clean & C & $\overline{\text{C}}$ & R & \multicolumn{2}{c}{ImageNet-P} & Clean & C & $\overline{\text{C}}$ & R & \multicolumn{2}{c}{Detection} \\
 & \textcolor{gray}{Error} & \textcolor{gray}{mCE} & \textcolor{gray}{Error} & \textcolor{gray}{Error} & \textcolor{gray}{mFR} & \textcolor{gray}{mT5D} &  \textcolor{gray}{RMS} & \textcolor{gray}{RMS} & \textcolor{gray}{RMS} & \textcolor{gray}{RMS} & \textcolor{gray}{AUROC ($\uparrow$)} & \textcolor{gray}{AUPR ($\uparrow$)}
\\
\hline
Baseline & 23.9&	78.2&	61.0& 63.8 &	58.0&	78.4&	5.6&	12.0&   20.7&	19.7&	79.7& 48.6 \\
Cutout & \underline{22.6} &	76.9&	60.2&	64.8 &  57.9&	75.2&	3.8&	11.1&	17.1&	14.6&	81.7& 49.6 \\
Mixup & 22.7&	72.7&	55.0&   62.3&	54.3&	73.2&	5.8&	7.3&	13.2&	44.6&	72.2& 51.3 \\
CutMix & 22.9&	77.8&	59.8&	66.5 & 60.3 &	76.6&	6.2&	9.1&	15.3&	43.5&	78.4& 47.9 \\
AutoAugment & \textbf{22.4}&	73.8&	58.0&	61.9 &  54.2&	72.0&	\textbf{3.6}&	8.0&	14.3&	12.6&	84.4& 58.2\\
AugMix & 22.8&	71.0&	56.5&   61.7&	52.7&	70.9&	4.5&	9.2&	15.0&	13.2&	84.2& 61.1 \\
SIN & 25.4&	70.9&	57.6&	\textbf{58.5}&   54.4&	71.8&	4.2&	6.5&	14.0&	16.2&	84.8& 62.3 \\
\textsc{PixMix} & \underline{22.6}&	\textbf{65.8}&	\textbf{44.3}&	\underline{60.1} & \textbf{51.1}&	\textbf{69.1}&	\textbf{3.6}&	\textbf{6.3}&	\textbf{5.8}&	\textbf{11.0}&	\textbf{85.7}& \textbf{64.1} \\
\Xhline{3\arrayrulewidth}
\end{tabular}
\caption{Full results for ImageNet. mT5D is an additional metric used for gauging prediction consistency in ImageNet-P. \textbf{Bold} is best, and \underline{underline} is second best.
}\label{tab:imagenet_appendix}
\end{table*}

\begin{table*}[ht]
\setlength\tabcolsep{5pt}
\small
\centering
\begin{tabular}{l | cccccccccccc } 
 & \multicolumn{1}{c}{Accuracy} & \multicolumn{3}{c}{Robustness} & \multicolumn{2}{c}{Consistency} & \multicolumn{4}{c}{Calibration} & \multicolumn{2}{c}{Anomaly} \\ \cmidrule(lr){2-2} \cmidrule(lr){3-5} \cmidrule(lr){6-7} \cmidrule(lr){8-11} \cmidrule(lr){12-13}
 & Clean & C & $\overline{\text{C}}$ & R & \multicolumn{2}{c}{ImageNet-P} & Clean & C & $\overline{\text{C}}$ & R & \multicolumn{2}{c}{Detection} \\
 & \textcolor{gray}{Error} & \textcolor{gray}{mCE} & \textcolor{gray}{Error} & \textcolor{gray}{Error} & \textcolor{gray}{mFR} & \textcolor{gray}{mT5D} &  \textcolor{gray}{RMS} & \textcolor{gray}{RMS} & \textcolor{gray}{RMS} & \textcolor{gray}{RMS} & \textcolor{gray}{AUROC ($\uparrow$)} & \textcolor{gray}{AUPR ($\uparrow$)}
\\
\hline
Baseline & 23.9&	78.2&	61.0& 63.8 &	58.0&	78.4&	5.6&	12.0&   20.7&	19.7&	79.7& 48.6 \\
Fractals & 	\textbf{22.0}&	68.2& 47.4 &60.6&52.6	&71.1	&4.0	&7.2	&  7.4 &11.7	&85.3	&62.6  \\
ResNet only FVis & 22.1&\textbf{64.3}	&  45.3&	\textbf{60.1}&  \textbf{50.7} &\textbf{69.1}	&3.9	&7.1	& 7.6  &12.2	&85.1	&63.3  \\
Fractals + FVis & 22.6&	65.8&	\textbf{44.3}&	\textbf{60.1} & 51.1&	\textbf{69.1}&	\textbf{3.6}&	\textbf{6.3}&	\textbf{5.8}&	\textbf{11.0}&	\textbf{85.7}& \textbf{64.1} \\
\Xhline{3\arrayrulewidth}
\end{tabular}
\caption{Similar to the results obtained in CIFAR-100 mixing set ablations, a fractal-only mixing set is effective (Fractals), but combining fractals and feature visualizations yields the best performance (Fractals + FVis). Moreover, feature visualizations from a model trained with the same dataset and architecture perform well (ResNet only FVis), showing that knowledge distillation does not explain the results.}\label{tab:imagenet_ablation}
\end{table*}

\begin{table*}[ht]
\setlength\tabcolsep{5pt}
\small
\centering
\begin{tabular}{l | cccccccccccc } 
 & \multicolumn{1}{c}{Accuracy} & \multicolumn{3}{c}{Robustness} & \multicolumn{2}{c}{Consistency} & \multicolumn{4}{c}{Calibration} & \multicolumn{2}{c}{Anomaly} \\ \cmidrule(lr){2-2} \cmidrule(lr){3-5} \cmidrule(lr){6-7} \cmidrule(lr){8-11} \cmidrule(lr){12-13}
 & Clean & C & $\overline{\text{C}}$ & R & \multicolumn{2}{c}{ImageNet-P} & Clean & C & $\overline{\text{C}}$ & R & \multicolumn{2}{c}{Detection} \\
 & \textcolor{gray}{Error} & \textcolor{gray}{mCE} & \textcolor{gray}{Error} & \textcolor{gray}{Error} & \textcolor{gray}{mFR} & \textcolor{gray}{mT5D} &  \textcolor{gray}{RMS} & \textcolor{gray}{RMS} & \textcolor{gray}{RMS} & \textcolor{gray}{RMS} & \textcolor{gray}{AUROC ($\uparrow$)} & \textcolor{gray}{AUPR ($\uparrow$)}
\\
\hline
Baseline & 23.9&	78.2&	61.0& 63.8 &	58.0&	78.4&	5.6&	12.0&   20.7&	19.7&	79.7& 48.6 \\
ANT & 23.9&	67.0&	61.0&	61.0&  48.0 &	68.4&	7.0&	10.3&	19.3&	22.9&	80.9& 54.3 \\
Speckle & 24.2&	72.7&	62.1&	62.1&   51.2&	70.6&	5.6&	11.6&	19.8&	20.9&	79.7& 53.3 \\
Noise2Net   & 22.7&	71.6&	57.7&	57.6&   51.5&	72.3&	4.4&	8.9&	16.3&	15.2&	84.8& 60.4 \\
EDSR   & 23.5&	65.4&	54.7&	60.3&   44.6&	63.3&	4.5&	8.4&	15.7&	16.7&	71.7& 36.3 \\
$\ell_\infty$ Adversarial & 45.5&	92.6&	68.0&	65.2&   38.5&	41.5&	15.5&	10.2&	15.1&	10.2&	69.8& 26.4 \\
$\ell_2$ Adversarial & 37.2&	85.5&	64.9&	63.0&   29.2&	34.8&	11.3&	9.7&	16.6&	10.7&	78.9& 40.2 \\
\Xhline{3\arrayrulewidth}
\end{tabular}
\caption{While many noise-based augmentation methods often do well on ImageNet-C by targeting the noise corruptions, they do not reliably improve performance across many safety metrics.}\label{tab:noise_methods}
\end{table*}

\begin{table*}[ht]
\footnotesize
\begin{center}
{\setlength\tabcolsep{5pt}%
\begin{tabular}{l | c | c  | c c | c c | c c | c c c c}
\multicolumn{3}{c}{} & \multicolumn{2}{c}{Noise} & \multicolumn{2}{c}{Blur} & \multicolumn{2}{c}{Weather} & \multicolumn{4}{c}{Digital} \\
\cline{2-13}
 & \multicolumn{1}{c|}{\,Clean\,} &{\,mFR\,} & \footnotesize{Gaussian}
    & \footnotesize{Shot} & \footnotesize{Motion} & \footnotesize{Zoom} & \footnotesize{Snow} & \footnotesize{Bright} & \footnotesize{Translate} & \footnotesize{Rotate} & \footnotesize{Tilt} & \footnotesize{Scale}\\ \hline 
Baseline            & 23.9 & 58.0 & 59	&58	&65	&72	&63	&62	&44	&52	&57	&48 \\
ANT       & 23.9 & 48.0 & 41	&36	&50&	61&	48&	58&	40&	48&	52&	46 \\
Speckle    & 24.2 & 51.2 & 38	&28	&60	&67	&58	&65	&43&	51	&54	&48 \\
Noise2Net           & 22.7 & 51.5 & 54	&53	&50&	70	&56	&50	&38	&47&	52&	43 \\
EDSR                 & 23.5 & 44.6 & 37	&35	&48	&56	&46	&56	&38	&44	&44	&43 \\
$\ell_\infty$ Adversarial       & 45.5 & 38.5 & 43	&56	&24&	33	&15	&80	&20	&34	&33	&46 \\
$\ell_2$ Adversarial       & 37.2 & 29.2 & 24	&30	&24	&31	&14	&64	&13	&27	&26	&39 \\
\Xhline{3\arrayrulewidth}
\end{tabular}}
\caption{ImageNet-P results. The mean flipping rate is the average of the flipping rates across all 10 perturbation types. Noise-based augmentation methods are less performant on non-noise distribution shifts.}
\label{tab:imagenetp}
\end{center}
\end{table*}

\begin{table*}[ht]
\setlength\tabcolsep{10pt}
\small
\centering
\begin{tabular}{l | ccc | ccc } 
 & \multicolumn{3}{c}{AUROC ($\uparrow$)} & \multicolumn{3}{c}{AUPR ($\uparrow$)} \\ \cmidrule(lr){2-4} \cmidrule(lr){5-7} 
 & Baseline & OE & \textsc{PixMix} & Baseline & OE & \textsc{PixMix} \\
\hline
Gaussian Noise &72.2	&93.5 &100.0	&23.5	&54.1 &100.0\\
Rademacher Noise &47.7	&90.2 &100.0	&14.6	&44.9 &100.0\\
Blobs &41.9	&100.0	&100.0	&13.0	&99.4 &100.0\\
Textures &66.6		&91.4	&80.3 &24.6		&75.7 &56.2\\
SVHN &96.6		&100.0	&99.5 &90.5		&99.9 &98.6\\
ImageNet &63.0		&86.5 &71.5	&25.1		&69.7 &47.4\\
Places69 &61.5		&63.1 &62.3	&23.4	&24.9	&31.3\\
\Xhline{\arrayrulewidth}
Average &64.2	&89.2 &87.6	&30.7	&66.9 &76.2\\
\Xhline{3\arrayrulewidth}
\end{tabular}
\caption{Out-of-Distribution detection results for a ResNet-18 pre-trained on Places365. \textsc{PixMix} and OE are finetuned for 10 epochs. Despite being a general data augmentation technique, \textsc{PixMix} is near the state-of-the-art in OOD detection.}\label{tab:appendix_anomaly_detection}
\end{table*}

\begin{table*}[ht]
\setlength\tabcolsep{9pt}
\small
\centering
\begin{tabular}{l | c | c | c } 
 & $k=2$ & $k=3$ & $k=4$ \\
\hline
$\beta=5$ & \makecell{20.2 \\ 31.6}	& \makecell{20.0 \\ 31.1} & \makecell{20.1 \\ 30.8} \\
\hline
$\beta=4$ & \makecell{19.7 \\ 31.3}	& \makecell{20.3 \\ 30.9} & \makecell{20.1 \\ 30.7}\\
\hline
$\beta=3$ & \makecell{20.3 \\ 31.2}	& \makecell{20.2 \\ 30.7} & \makecell{20.3 \\ 30.5}\\
\Xhline{3\arrayrulewidth}
\end{tabular}
\caption{Performance is not strongly affected by hyperparameters. We include the CIFAR-100 test set error and the CIFAR-100-C mCE for each hyperparameter setting.}\label{tab:cifar_appendix5}
\end{table*}

\begin{table*}[ht]
\footnotesize
\begin{center}
{\setlength\tabcolsep{3pt}%
\begin{tabular}{l | c | c | c c c | c c c c | c c c  c | c c c c}
\multicolumn{3}{c}{} & \multicolumn{3}{c}{Noise} & \multicolumn{4}{c}{Blur} & \multicolumn{4}{c}{Weather} & \multicolumn{4}{c}{Digital} \\
\cline{2-18}
 & \multicolumn{1}{c|}{\,Clean\,} &{\,mCE\,} & \scriptsize{Gauss.}
    & \scriptsize{Shot} & \scriptsize{Impulse} & \scriptsize{Defocus} & \scriptsize{Glass} & \scriptsize{Motion} & \scriptsize{Zoom} & \scriptsize{Snow} & \scriptsize{Frost} & \scriptsize{Fog} & \scriptsize{Bright} & \scriptsize{Contrast} & \scriptsize{Elastic} & \scriptsize{Pixel} & \scriptsize{JPEG}\\ \hline 
Baseline & 23.9 & 78.2 & 78 & 80 & 80 & 79 & 90 & 81 & 80 & 80 & 78 & 69 & 62 & 75 & 88 & 76 & 78 \\
Cutout & 22.6 & 76.9 & 76 & 77 & 79 & 76 & 90 & 79 & 79 & 79 & 78 & 69 & 60 & 74 & 87 & 75 & 75 \\
Mixup & 22.7 & 72.7 & 69 & 72 & 73 & 76 & 90 & 77 & 78 & 73 & 68 & 62 & 59 & 64 & 86 & 71 & 73 \\
CutMix & 22.9 & 77.8 & 78 & 80 & 80 & 79 & 90 & 81 & 80 & 80 & 78 & 69 & 62 & 75 & 88 & 76 & 78 \\
AutoAugment & 22.4 & 73.8 & 71 & 72 & 75 & 75 & 90 & 78 & 79 & 73 & 74 & 64 & 55 & 68 & 87 & 73 & 71 \\
AugMix & 22.8 & 71.0 & 69 & 70 & 70 & 72 & 88 & 74 & 71 & 73 & 74 & 58 & 58 & 59 & 85 & 73 & 72 \\
SIN & 25.4 & 70.9 & 64 & 65 & 66 & 73 & 84 & 73 & 80 & 71 & 74 & 66 & 62 & 69 & 80 & 64 & 73 \\
\textsc{PixMix} & 22.6 & \textbf{65.8} & 53 & 52 & 51 & 73 & 88 & 77 & 77 & 62 & 64 & 58 & 56 & 53 & 85 & 69 & 70 \\
\Xhline{3\arrayrulewidth}
\end{tabular}}
\caption{Clean Error, mCE, and Corruption Error (CE) values for various methods on ImageNet-C. The mCE value is computed by averaging across per corruption CE values.
}
\label{tab:imagenet-table}
\end{center}
\end{table*}

\begin{table*}[ht]
\footnotesize
\begin{center}
{\setlength\tabcolsep{3.5pt}%
\begin{tabular}{l | c | c | c c c c c c c c c c}
 & \multicolumn{1}{c|}{\,Clean\,} &{\,$\overline{\text{C}}$ Error\,} & \scriptsize{Blue Sample}
    & \scriptsize{Plasma} & \scriptsize{Checkerboard} & \scriptsize{Cocentric Sine} & \scriptsize{Single Freq} & \scriptsize{Brown} & \scriptsize{Perlin} & \scriptsize{Sparkles} & \scriptsize{Inverse Sparkle} & \scriptsize{Refraction}\\ \hline 
Baseline & 23.9 & 61.0 & 62 & 77 & 55 & 86 & 80 & 45 & 41 & 38 & 78 & 48 \\
Cutout & 22.6 & 60.2 & 64 & 77 & 49 & 85 & 80 & 45 & 41 & 36 & 77 & 47 \\
Mixup & 22.7 & 55.0 & 58 & 68 & 49 & 80 & 72 & 38 & 36 & 35 & 71 & \textbf{44} \\
CutMix & 22.9 & 59.8 & 64 & 77 & \textbf{47} & 85 & 80 & 46 & 41 & 35 & 75 & 47 \\
AutoAugment & 22.4 & 58.0 & 56 & 71 & 49 & 86 & 77 & 42 & 39 & 36 & 77 & 47 \\
AugMix & 22.8 & 56.5 & 51 & 71 & 48 & 83 & 76 & 42 & 38 & 36 & 75 & 45 \\
SIN & 25.4 & 57.6 & 53 & 72 & 54 & 81 & 68 & 41 & 41 & 41 & 79 & 47 \\
\textsc{PixMix} & 22.6 & \textbf{44.3} & \textbf{40} & \textbf{48} & 48 & \textbf{48} & \textbf{47} & \textbf{34} & \textbf{37} & \textbf{33} & \textbf{65} & \textbf{44} \\
\Xhline{3\arrayrulewidth}
\end{tabular}}
\caption{Results for various methods on ImageNet-$\overline{\text{C}}$.}
\label{tab:imagenetcbar}
\end{center}
\end{table*}